\pdfoutput=1

\documentclass[11pt]{article}

\usepackage{amsmath}

\usepackage{ACL2023}

\usepackage{times}
\usepackage{latexsym}

\usepackage[T1]{fontenc}

\usepackage[utf8]{inputenc}

\usepackage{microtype}

\usepackage{inconsolata}

\usepackage{amsfonts}
\usepackage{graphicx}
\usepackage{booktabs}
\usepackage{adjustbox}

\usepackage{color, colortbl}

\usepackage{arydshln} %
\usepackage{algorithm}
\usepackage{algpseudocode}
\usepackage{multirow}
\usepackage{subcaption}
\usepackage{float}

\usepackage{todonotes}

\title{Balancing the Scales: Reinforcement Learning for Fair Classification}

\author{Leon Eshuijs \\
  Vrije Universiteit Amsterdam  \\ Amsterdam, the Netherlands \\
  \texttt{l.eshuijs@vu.nl} \\\And
    Shihan Wang \\
  Utrecht University  \\ Utrecht, the Netherlands \\
  \texttt{s.wang2@uu.nl} \\\And
  Antske Fokkens \\
  Vrije Universiteit Amsterdam  \\ Amsterdam, the Netherlands \\
  \texttt{antske.fokkens@vu.nl}   }

\begin{document}

\maketitle
\begin{abstract}
Fairness in classification tasks has traditionally focused on bias removal from neural representations, but recent trends favor algorithmic methods that embed fairness into the training process. These methods steer models towards fair performance, preventing potential elimination of valuable information that arises from representation manipulation. Reinforcement Learning (RL), with its capacity for learning through interaction and adjusting reward functions to encourage desired behaviors, emerges as a promising tool in this domain. In this paper, we explore the usage of RL to address bias in imbalanced classification by scaling the reward function to mitigate bias. We employ the contextual multi-armed bandit framework and adapt three popular RL algorithms to suit our objectives, demonstrating a novel approach to mitigating bias.
\footnote{Our code is available at \url{https://github.com/watermeleon/RL_for_imbalanced_classification}}

\end{abstract}

\section{Introduction}
In recent years, the issue of bias and fairness in Artificial Intelligence and Natural Language Processing has received significant attention \cite{mehrabi2021survey}. 
In decision-making models such as classification algorithms, bias 
often stems directly from the training data leading to unfair outcomes between protected groups such as gender or race.
To address this problem, previous work on fairness has focused on achieving \textit{representational fairness}, so that the information of the protected groups is lost
\cite{ravfogel2020null, haghighatkhah2022better}.
However, recent work has demonstrated no meaningful correlation between representational fairness and \textit{empirical fairness}, i.e. fairness on downstream tasks \cite{shen2022does}.
To address empirical fairness directly, other work has explored the intersection of bias mitigation and class-imbalanced learning \cite{subramanian2021fairness}. 
Class-imbalanced learning approaches aim to achieve fair performance by balancing the training data via sampling or reweighing the loss function.

\begin{figure}[t]
     \centering
     \includegraphics[width=1.0\columnwidth]{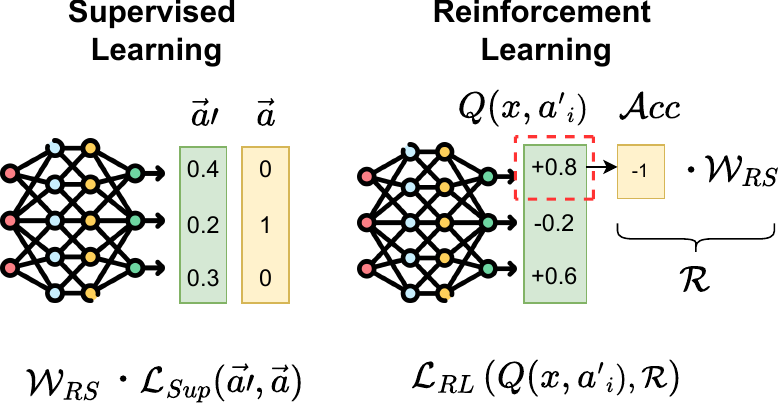}
    \caption{Overview of the classification setup with input vector $x$, and output class $a$ for Reinforcement Learning and Supervised Learning, highlighting the place of the reward scaling matrix $\mathcal{W}_{RS}$ }
    \label{fig:algo_Sup_RL_sketch}
\end{figure}

At the same time, Reinforcement Learning (RL) has emerged as a promising alternative to traditional supervised learning methods for various NLP tasks, including syntactic parsing, conversational systems, and machine translation \cite{uc2023survey}. 
Unlike traditional supervised learning methods, RL is not bound to binary labels and is trained directly on the continuous value of each input, as illustrated in Figure \ref{fig:algo_Sup_RL_sketch}. 
RL agents can learn from sparse reward signals, receiving the rewards for the action they choose, not necessarily the correct one.
By exploring the environment and adapting their behavior based on the received rewards, RL agents find optimal actions under varying state values.
In the context of classification, RL has been adapted to mitigate class imbalance through a scaling component of the reward function for binary classification \cite{lin2020deep}.
However, implementations considering more complex imbalances have remained largely unexplored.

In this work, we leverage RL to address fairness among protected groups in multi-class classification. First, we propose to frame the fair classification task as a Contextual Multi-Armed Bandit (CMAB) problem, see Figure \ref{fig:algo_Sup_RL_sketch} for an overview of our setup.
To mitigate bias, we scale the reward function to counteract imbalances among protected groups within each class.
We employ three different types of RL methods, each reflecting a key type of RL approach, and adapt them for our task.
Additionally, we integrate the different scaling approaches into a supervised learning baseline to evaluate the effectiveness of our RL-based methods.

Experiments on two fair classification datasets demonstrate that our RL algorithms achieve competitive performance compared to existing baselines and that reward scaling is a powerful tool to mitigate bias in classification.
We further investigate how stable reward scaling is under various class and subclass imbalances as well as various degrees of \textit{representational fairness}.
Notably, the deep RL algorithms perform best on the multi-class dataset, while the classical CMAB algorithm excels on the binary dataset.
Moreover, our scaled supervised implementation exceeds existing implementations and shows state-of-the-art performance for multi-class classification.

\section{Related work}
\subsection{Bias Mitigation}

Research on mitigating bias can be divided in those that tackle the training data \cite{wang2019balanced}, those that attempt to remove bias from representations \cite{ravfogel2020null, haghighatkhah2022better}, and those that adjust the learning process \cite{elazar2018adversarialEmoji,han2021diverse}. 
Within approaches that adjust the learning processes, we distinguish two main categories: those that add adversarial learners to ignore protected attributes \cite{wadsworth2018achieving}, and more closely to our work, approaches that adjust the loss function to emphasize performance on minority classes. 

Prior work that modified the training setup to increase fairness used methods such as down/upsampling \cite{wang2019balanced} and reweighting the loss function \cite{hofler2005use, lahoti2020fairness}.
\newcite{han2022balancing} evaluate both down-sampling and loss reweighting on two datasets for fair text classification. Both techniques are applied to align training with different definitions of fairness. Downsampling using the Equal Opportunity fairness metric demonstrated impressive results. In this paper, we take the first step to explore whether reward scaling in reinforcement learning can improve fairness in classification.

\subsection{Reinforcement Learning for Classification}
Literature on RL applications for classification predominantly considers the following two theoretical frameworks: Markov Decision Process (MDP) and the Contextual Multi-Armed Bandit (CMAB).

Early work by \newcite{wiering2011reinforcement} casts classification as a sequential decision-making task, by introducing a classification variant of the MDP.
In their setup agents manipulate memory cells to encode information by applying an action sequence on a single sample. They demonstrated competitive performance, but, remained limited to small tasks due to the computational complexity.
\newcite{lin2020deep} extended this work, by introducing a variant of the classification MDP and applying a Deep Q-learning Network (DQN) to binary classification of images and texts.
They focused on mitigating bias arising from class imbalance by scaling the rewards inversely proportional to the class frequency. 
However, in their setup the sequential component was taken over multiple data points, which assumes sequential dependency among data samples in the classification task.

The RL framework CMAB offers a promising alternative because it considers the input as a sequence of independent states. We formalize our classification task as a CMAB problem, because this is consistent with the independence of data points in the commonly shuffled datasets. 
\newcite{dudik2014doubly} use CMAB agents by modifying K-class classification as a K-armed bandit problem, where the agent receives a reward of 1 for correct and 0 for incorrect classification. \newcite{dimakopoulou2019balanced} use this framework and modify different CMAB algorithms to balance exploration and exploitation and compare the original and modified agents on 300 classification datasets. However, their analysis focused on datasets with either limited classes, features, or observations. 
To the best of our knowledge, we are the first to extend reward scaling for fair multi-class classification or to apply reward scaling for classification with CMAB.

\section{Methodology}

In this section, we describe how we formalize our classification task as a CMAB. We introduce three RL methods and explain how we adapt them for fair classification.\footnote{Due to space limitations, we only summarize the key idea of the algorithms and how we adapt them in the paper. Please refer to the Appendix and original papers for more details.}

\subsection{Contextual Multi-Armed Bandit} \label{sec:cmab}
We formalize the multi-class classification task as a finite contextual multi-armed bandit (CMAB) problem. 
In each round $t$, an agent is presented with a context vector $x_t \in \mathbb{R}^d$. 
The agent chooses an action $a_t \in A$ from a fixed set of arms, based on the policy $a_t \sim \pi(x_t)$. After the action is taken, the environment returns a reward: $r_t \sim \mathcal{R}$. 
In a multi-class classification framework, the action space is the set of all possible classes, while the context vector is a representation of the input, e.g.\ a contextual text embedding (see Section \ref{lab:contex_embedding} for more information). 
Within a finite number of rounds, the agent aims to learn the optimal policy to maximize the total reward. In other words, given a set of testing data, we aim to learn the optimal policy to maximize the selection of correct classes. 

We extend the CMAB framework for fair classification by constructing a reward function that counters data imbalances. We assign a reward scale for each sensitive state  $(a,g)$, comprising the desired class $a$ (e.g.\ occupation) and protected attribute $g$ (e.g.\ gender).
The total reward for a given prediction is calculated as $\mathcal{R}(a, a_{pred}, g) = \mathcal{A}cc(a, a_{pred}) \cdot \mathcal{W}(a, g)$. It comprises an accuracy term $\mathcal{A}cc$, and a reward scale matrix $\mathcal{W}$. 
Unlike previous work \cite{dudik2014doubly}, which defines the accuracy term as $\mathcal{A}cc \in \{0,1\}$, we define it as $\mathcal{A}cc \in \{-1,1\}$. Which allows us to scale the reward for both correct (+1) and incorrect classifications (-1). We use the term \textit{reward scale} to indicate that this approach adjusts the magnitude but not the sign of the reward. 
Section~\ref{sec:reward_scales} presents various designs of the reward scale.

\subsection{Reinforcement Learning Algorithms}

We select three different RL algorithms and adapt them to learn optimal policies for fair classification in the formalized CMAB problem. These algorithms include one classical CMAB algorithm that addresses the linear relationship between the expected reward and the context, as well as two popular deep RL algorithms for MDP problems, Deep Q-Network (DQN) and Proximal Policy Optimization (PPO), which allow us to leverage non-linear approximations. The two deep RL algorithms are selected as they are representative of the two key types of deep RL approaches: value-based methods and policy gradient methods. By employing these three algorithms, we aim to investigate the application of diverse RL methods. 

\subsubsection{LinUCB}
The classical CMAB algorithm, disjoint  Linear UCB (LinUCB)  \cite{linucb_li2010contextual} assumes a linear relationship between the context embedding $x_t$ and the reward $E[r_{t, a}|x_{t}] = x_{t}^\top\theta_a$. A benefit of disjoint LinUCB over other CMAB algorithms is that each class has a unique learnable weight vector $\theta_a$, which makes it suitable for classification with many classes.
In each round, the agent chooses the arm (i.e.\ class label) with the highest score $\hat{\theta}_a^\top  x_{t} + \alpha \sqrt{x_{t}^\top A_a^{-1} x_{t}}$, based on the context vector $x_t$. This is a combination of the mean of the expected payoff,  $ \hat{\theta}_a^\top  x_{t}$, and the standard deviation $\sqrt{x_{t}^\top A_a^{-1} x_{t}} $, weighted with parameter $\alpha$ to control the level of exploration.
The weight vector of each arm is defined as $\hat{\theta}_{a_t} = A_{a_t}^{-1}b_{a_t}$. Here the covariant matrix $A_{a_t}$ is calculated with the history of context vectors chosen by that arm,  $A_a = \lambda I_d + \sum_{s=1}^{t-1} x_{s} x_{s}^\top$. The vector $b_{a}$ is the mean context vector of the arm weighted by the obtained rewards, $b_{a_t} = \sum_{s=1}^{t} r_{s,a_t} x_{s,a_t}$. 

\subsubsection{$\textbf{DQN}_{\textbf{bandit}}$}\label{sec:DQN_subsection}
To adapt the MDP algorithms for a CMAB problem, our CMAB implementation is congruent with a one-step MDP, where each initial state is sampled from the existing set of context $s_1 \in  X$, and each second state is a terminal state.
In DQN \cite{mnih2015humanDQN}, the agent learns a Q-function, parameterized by $\phi$, to estimate the return for each state-action pair.
According to the Bellman equation \cite{Bellman1957}, the optimal Q-value, $Q^*$, of two sequential states are linked by: 
\begin{equation}\label{eq:DQN_bellman}
    Q^*(s, a)=\mathbb{E}_\pi [r_t+\gamma  \max _{a'} Q^*\left(s_{t+1}, a' \right) ]
\end{equation}

In our case (the one-step MDP), each next state is the terminal state, after which there is no reward, thus we obtain, $Q_{\phi}\left(s_{t+1}, \cdot\right) = 0$, and $G_t = r_t$.
The parameters of $\phi$ are optimized using the mean-squared error between the current Q-value, $Q_{\phi}(s_{t}, a_{t})$, and the updated value provided in Equation~\ref{eq:DQN_bellman}. 
The updated value is computed as $r_{t}+ \gamma \max _{a '} Q_{\phi}\left(s_{t+1}, a' \right)$, but since the next state is always the terminal state it reduces to  $r_{t}$.
We finalize the adaptation of DQN for the CMAB by casting the states as context vectors, obtaining the loss function:
\begin{equation*}
    L_{DQN}(\phi) = \mathbb{E}_{(x_t, a_t, r) \in B} \left[ ( r - Q_{\phi}(x_t, a_t))^2 \right]
\end{equation*}
The network is updated by sampling a minibatch of tuples $B$ from the replay buffer. The $DQN_{bandit}$ enables exploration using an $\epsilon$-greedy policy for selecting actions.

\subsubsection{$\textbf{PPO}_{\textbf{bandit}}$}

Different from DQN, in Proximal Policy Optimization (PPO) \cite{schulman2017proximal}, the policy $\pi$ (parameterized by $\theta$) is directly optimized under the objective of selecting the best action.
The general objective in policy gradient methods is to maximize:
$\mathbb{E}_{\tau \sim \pi_{\theta}} \left[ \sum_{t=0}^{T} \nabla_{\theta} \log \pi_{\theta}(a_t|s_t) \cdot A_t \right]$. The advantage $A_t$ is computed as $A_t = Q_\pi(s_t, a_t) - V_{\phi}(s_t)$, where a critic network $V_\phi$ is used to estimate the state value. 
PPO ensures the policy does not deviate too far during an update, by scaling the advantage with the probability ratio, $r_t(\theta)$.
This ratio is clipped to create a conservative lower bound to control the policy's change at each step.
The actor's objective function is thus defined as:
\begin{equation*}
 L_{actor}(\theta) = \mathbb{E}_t \left[ \min(r_t(\theta) A_t, \text{clip}_\epsilon(r_t(\theta)) A_t) \right] 
\end{equation*}

To adapt PPO for CMAB, the sequential component is removed and the state $s_t$ is replaced by the context vector $x_t$. For the actor loss, the advantage changes and is calculated as $A_t = r_t - V_{\phi}(x_t)$.
The return again reduces to the reward, thereby simplifying the critic loss to:
\begin{equation*}
    L_{critic}(\phi) = \mathbb{E}_t \left[ (V_{\phi}(x_t) - r_t)^2 \right]
\end{equation*}
Lastly, the final loss of the $PPO_{bandit}$ agent contains a penalty that maximizes the policy's entropy of the context vector to encourage exploration.

\subsection{Reward Scales} \label{sec:reward_scales}
\begin{figure}[t]
     \centering
     \includegraphics[width=0.95\columnwidth]{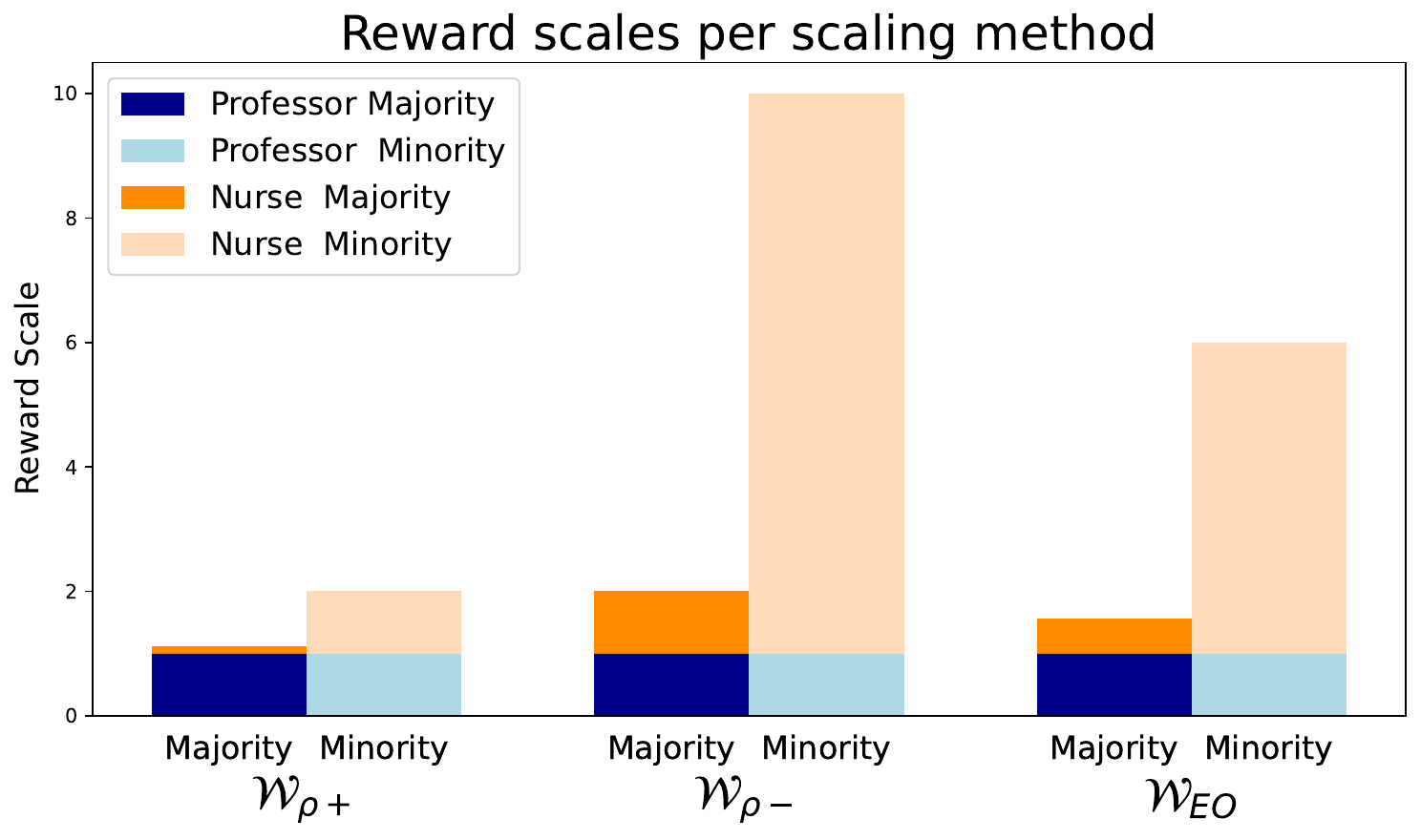}
    \caption{Reward scales for the professions with different gender imbalances \textit{Professor} (50/50) and \textit{Nurse} (90/10) using the different scaling functions. }
    \label{fig:reward_scales_example}
\end{figure}

 \newcite{lin2020deep} implement reward scaling on an imbalanced binary classification dataset.
Inspired by this, we propose different ways of reward scales for multi-class classification with imbalances of protected attributes. 
For context, we use the profession classification dataset, BiasBios, where reward scaling tackles the sub-class imbalance of the protected attribute gender.
To illustrate the influence of various reward scales Figure \ref{fig:reward_scales_example} shows the scales of a balanced (Professor) and an imbalanced (Nurse) class for the protected groups with attribute gender.

For the first method, we extend the implementation of \newcite{lin2020deep} into the multi-classification setting and reduce the reward for the majority by scaling it with the imbalanced ratio $\rho^a_{imb} = \frac{|D^a_{min}|}{|D^a_{maj}|}$, which is the ratio between the number of samples of the minority and majority class in class $a$. 
$$
\mathcal{W}_{\rho +}(a, g) = \begin{cases} 
      1 & \text{if } g \text{ is minority in } a \\
      \rho^a_{imb} & \text{if }  g \text{ is majority in } a 
   \end{cases}
$$

Figure \ref{fig:reward_scales_example} demonstrates that $\mathcal{W}_{\rho +}(x)$ scales with a reverse of the bias within a class, however, compared to a balanced class, the reward scale of the majority is very low.
Therefore, we propose a second design that keeps the scales of the majority group in the imbalanced class equal to the scales of the balanced class.
Thus we set the majority value at 1 and only increase the minority value, based on the inversed imbalanced ratio. 
$$
\mathcal{W}_{\rho -}(a, g) = \begin{cases} 
      (\rho^a_{imb})^{-1}  & \text{if } g \text{ is minority in } a \\
      1 & \text{if }  g \text{ is majority in } a 
   \end{cases}
$$

The third implementation adopts the Equal Opportunity (EO) formalization used by \newcite{han2022balancing}. Contrary to the previous two methods it ensures the average weights per class remain equal, providing an improved theoretical fairness among classes.
The EO objective is achieved by aggregating the loss per sensitive state and then scaling it.
However, our work scales per instance, thus we convert the EO objective to instance-specific weights (see Appendix \ref{sec:eo_appendix}) and obtain: 
$$ \mathcal{W}_{EO}(a, g) = \frac{1}{2} \frac{1}{P(g|a)} $$

Lastly, we also employ the Inverse Probability Weighting (IPW) technique \cite{hofler2005use}. Full fairness across classes and protected groups is obtained by scaling with the joint probability, resulting in:
$$
\mathcal{W}^{IPW}(a, g) =  \frac{1}{P(a, g)}$$

\subsection{Supervised Learning: Loss reweighting}
\label{sup}
Parallel to reward scaling in RL is (instance) reweighing in supervised learning \cite{han2022balancing, lahoti2020fairness}, here loss reweighting for clarity.
Loss reweighting has been a popular technique for imbalanced datasets, where the loss of each data sample is scaled to mitigate the class imbalance, traditionally using the IPW \cite{hofler2005use}. 
The weighted cross-entropy loss using the true probability $p$, predicted probability $q$:
\begin{equation*}
L^{CE}= -\sum_{x,g} \sum_{a}  \mathcal{W}(a, g) p(a|x) \log q(a|x) 
\end{equation*}

We implement supervised learning with loss reweighing to serve as a strong baseline and highlight the connection between loss reweighing and reward scaling.

\definecolor{Gray}{gray}{0.9}

\section{Experiments}
\subsection{Dataset}
The BiasBios \cite{de2019bias} consists of 393,423 biographies labeled with one of 28 professions, and a binary gender label. 
Following \newcite{de2019bias}, the data is randomly split according to 65\% training, 25\% testing, and 10\% for validating. The dataset contains two imbalances: 
varying frequencies of the professions and a difference in gender percentage for each class.

Following \newcite{ravfogel2020null} and \newcite{han2022balancing} we also evaluate on the Emoji \citep{elazar2018adversarialEmoji} sentiment analysis task of Twitter data \cite{blodgett2016demographicEmoji}. 
The task involves binary sentiment classification evaluation with race as the protected attribute, approximated through the provided labels Standard American English (SAE) and African American English (AAE). 
As per \newcite{han2021diverse}, the dataset is composed of Happy (40\% AAE, 10\% SAE), and Sad: (10\% AAE, 40\% SAE). We use the same train, dev, and test splits of 100k/8k/8k instances, respectively.

\paragraph{Context Vectors} \label{lab:contex_embedding}
Each textual data sample is embedded into a context vector via a pretrained encoder, enabling the algorithms for classification.
Following \newcite{ravfogel2020null} we use the same fixed pretrained encoder for each dataset. For the BiasBios dataset, each biography is encoded using the \texttt{[CLS]} output of the uncased BERT-base model \cite{devlin2018bert}.
For the Emoji dataset, we use the DeepMoji encoder \cite{felbo2017using}, which has been demonstrated to capture a diverse range of moods and demographic information. 

\subsection{Metrics}
Following prior work, we evaluate performance using accuracy and fairness using the True Positive Ratio (TPR) gap \cite{de2019bias,ravfogel2020null}. 
The TPR gap of a class $a \in A$ is calculated as:
$TPR_{gap}^a = TPR_g^a - TPR_{\sim g}^a $, where $g$ and $\sim g$ represent the two options for the sensitive states.
The global TPR metric, GAP, is then calculated as the root mean square of the individual metrics:
\begin{equation}
    GAP=\sqrt{\frac{1}{|A|} \sum_{a \in A }\left(TPR^{a}_{gap}\right)^2}
\end{equation}

To quantify performance and fairness as a single metric we use the \textit{Distance To the Optimum} (DTO) introduced in \newcite{han2022balancing}. 
DTO combines the metrics (accuracy, 1-GAP) as dimensions of evaluation space and computes the Euclidean distance between the achieved and Utopian point. The smaller the distance to the Utopian point (lower DTO), the better. We report the DTO with the Utopian accuracy and GAP as the best values across all evaluated models.

While accuracy measures the overall performance and GAP the disparity among protected groups within a class, these metrics do not capture imbalance performance across classes. Therefore we also evaluate our algorithms using the macro-averaged F1 metric to detect if minority classes are ignored. 
All metrics are scaled by 100 for ease of reading and all metrics are represented in the tables as the mean $\pm$ std over 5 random seeds, except DTO which is taken over the mean.

\subsection{Hyperparameters and model selection}

Each algorithm uses the same classifier architecture, except LinUCB, which has a custom set of learnable parameters. The classifier has one hidden layer MLP. 
All models are trained for 10 epochs, except LinUCB, which achieved optimal performance within 2 epochs.
All models are evaluated on the validation set after 50k iterations to account for different convergence speeds of models.
The best model throughout training and across hyperparameters is selected using DTO. We apply hyperparameter optimization on both datasets for each of the algorithms, for details see Appendix \ref{sec:hyperparam_tuning}.

\begin{table}[t]
\centering
\resizebox{\columnwidth}{!}{
\begin{tabular}{@{}lcccc@{}}
       & \multicolumn{2}{c}{\textbf{PPO}$_{bandit}$} & \multicolumn{2}{c}{\textbf{Sup}} \\ 
        \cmidrule(lr){2-3} \cmidrule(lr){4-5}
       & \textbf{Accuracy} \small{$\uparrow$}& \textbf{GAP} \small{$\downarrow$}& \textbf{Accuracy} \small{$\uparrow$}& \textbf{GAP} \small{$\downarrow$}\\ \midrule
\textbf{$\mathcal{W}_{\rho +}$} & 74.6 $\pm$ 0.7& 9.9 $\pm$ 0.8 & 79.3 $\pm$ 0.1& 7.9 $\pm$ 0.3 \\
\textbf{$\mathcal{W}_{\rho -}$} & 78.8 $\pm$ 0.1& \textbf{8.4 $\pm$ 0.6} & 79.8 $\pm$ 0.3& 6.9 $\pm$ 0.2 \\
\textbf{$\mathcal{W}_{EO}$}     & \textbf{79.2 $\pm$ 0.2}& 8.5 $\pm$ 0.2 & \textbf{80.1 $\pm$ 0.2}& 7.1 $\pm$ 0.5 \\ 
\textbf{$\mathcal{W}_{IPW}$} & 45.8 $\pm$ 6.9& 10.5 $\pm$ 0.9 & 72.1 $\pm$ 0.7& \textbf{6.1 $\pm$ 0.3} \\ \bottomrule
\end{tabular}}
\caption{Results with different reward scales for Supervised Learning (Sup) and PPO on BiasBios}
\label{tab:reward_scale_metric_results}
\end{table}

\begin{table*}[t!]
\centering
\resizebox{\textwidth}{!}{
\begin{tabular}{@{}llrclcllcc}
 & \multicolumn{5}{c}{\textbf{BiasBios} (28 Classes)} &   \multicolumn{4}{c}{\textbf{Emoji} (2 Classes)}\\
\multicolumn{1}{c}{\textbf{Algorithm}}  & \multicolumn{1}{c}{\textbf{Accuracy} \small{$\uparrow$}}  & \multicolumn{1}{c}{\textbf{GAP} \small{$\downarrow$}} & \multicolumn{1}{c}{\textbf{DTO} \small{$\downarrow$}} & \multicolumn{1}{c}{\textbf{F1} \small{$\uparrow$}} &\multicolumn{1}{c}{\textbf{Time} \small{$\downarrow$}} & \multicolumn{1}{c}{\textbf{Accuracy} \small{$\uparrow$}}  & \multicolumn{1}{c}{\textbf{GAP} \small{$\downarrow$}} & \multicolumn{1}{c}{\textbf{DTO} \small{$\downarrow$}}&\multicolumn{1}{c}{\textbf{Time} \small{$\downarrow$}} \\ \midrule
Sup & 81.0 $\pm$ 0.1& 16.4 $\pm$ 0.5& 9.3& 73.8 $\pm$ 0.3 & 1.0 & 72.3 $\pm$ 0.1& 38.1 $\pm$ 0.6& 28.3& 1.0\\
INLP   &  80.2 $\pm$ 0.6& 9.7 $\pm$ 0.4&  2.8&  71.7 $\pm$ 1.4 & 50.1& 63.5 $\pm$ 3.6&  24.1 $\pm$ 5.4& 18.6& 3.6\\
MP& \textbf{81.1 $\pm$ 0.1} & 13.9 $\pm$ 0.6& 6.8& \textbf{74.0 $\pm$ 0.2}& 2.6& 71.8 $\pm$ 0.3& 17.1 $\pm$ 1.0& 8.1&2.3\\
BTEO& 79.2 $\pm$ 0.3&    8.4 $\pm$ 0.6& 2.3& 68.1 $\pm$ 0.4& 1.7& 75.4 $\pm$ 0.1& 10.4 $\pm$ 1.0 & \textbf{0.4} & 0.8\\
DAdv& 80.8 $\pm$ 0.2&  8.5 $\pm$ 0.6&  1.4&72.9 $\pm$ 0.4& 4.8&\textbf{75.6 $\pm$ 0.3}& 11.6 $\pm$ 1.7& 1.6& 5.7\\
\rowcolor{Gray}
Sup$^{ {EO}}$     & 80.1 $\pm$ 0.2& \textbf{7.1 $\pm$ 0.5}& \textbf{1.0}& 71.7 $\pm$ 0.5 &1.0& 75.5 $\pm$ 0.1 & 11.4 $\pm$ 1.1 & 1.4&1.0\\
\rowcolor{Gray}
LinUCB$^{ {EO}}$  & 74.6 $\pm$ 0.2& 12.2 $\pm$ 0.5& 8.3& 59.8 $\pm$ 1.1 &31.9& 75.3 $\pm$ 0.2& 10.4 $\pm$ 0.7  & 0.5& 2.8 \\
\rowcolor{Gray}
$\text{DQN}_{bandit}^{ {EO}}$ & 79.2 $\pm$ 0.1& 10.1 $\pm$ 0.4& 3.6&66.4 $\pm$ 0.2 &57.4& 70.8 $\pm$ 0.8& \textbf{10.0 $\pm$ 1.0} &4.8&30.2\\ 
 \rowcolor{Gray}
$\text{PPO}_{bandit}^{{EO}}$  & 79.2 $\pm$ 0.2& 8.5 $\pm$ 0.2& 2.4&66.0 $\pm$ 0.8 &2.9& 75.4 $\pm$ 0.1& 14.4 $\pm$ 0.6  & 4.4&3.0\\  \bottomrule
\end{tabular}
}
\caption{Results on the BiasBios and Emojis classification datasets for our own models (in grey) and the baselines. Metrics are provided as mean $\pm$ std over 5 random seeds, except DTO which is computed over the mean Accuracy, and GAP, and Time which is the relative time compared to the supervised baseline (first row). }
\label{tab:mega_table_baselines}
\end{table*}

\subsection{Comparison Models}
Besides the supervised implementation in Section~\ref{sup}, abbreviated to \textbf{Sup}, we also compare our models against various existing debiasing methods.
\textbf{INLP} \cite{ravfogel2020null} debiases embeddings by iteratively training classifiers to predict the protected attribute, it then removes this information from the embedding using a projection of the classifier's nullspace.
\textbf{MP} \cite{haghighatkhah2022better} simplifies the INLP setup by using a single Mean Projection (MP) between the representation of each class's protected groups.
\textbf{DAdv}, \cite{han2021diverse} removes sensitive information from the embeddings by applying adversarial training using diverse adversaries. Lastly, \textbf{BTEO} \cite{han2022balancing} subsamples the dataset to establish equal opportunity.
We implement these existing methods with the same training settings for fair comparison. Notably, we highlight how Supervised $\mathcal{W}^{EO}$ is theoretically equal to instance reweighing in \newcite{han2022balancing}, but our implementation achieves significantly higher performance.

\section{Results and Analysis}
We train each of the RL algorithms with and without reward scaling. As a strong baseline, we also train a supervised learning model in the standard fashion and use loss reweighting with the same reward scale function.

\subsection{Different Reward Functions}
We first evaluate the effect of the different reward scales discussed in Section~\ref{sec:reward_scales} by providing the results for the implementations of PPO and Sup, see Table~\ref{tab:reward_scale_metric_results}, for full table see Appendix~\ref{sec:appendix_4_4_reward_scales}.

The results presented in the table demonstrate that the imbalance ratio $\rho$ yields substantial gains in fairness and accuracy when applied to increase the reward for the minority class (\textbf{$\mathcal{W}_{\rho -}$}) as opposed to decreasing the reward for the majority class (\textbf{$\mathcal{W}_{\rho +}$}).
Especially the accuracy of PPO is sensitive to this, suggesting that PPO might not work very well for states with low reward scales.

As hypothesized, scaling with the joint probability of class and protected attribute as in \textbf{$\mathcal{W}_{IPW}$} proves to be too unstable. It results in the worst accuracy for both algorithms, with a minor improvement in fairness for supervised learning. The overall difference between \textbf{$\mathcal{W}_{EO}$} and \textbf{$\mathcal{W}_{\rho -}$} is minimal as expected from their similar reward scales depicted in Figure~\ref{fig:reward_scales_example}. 
We use EO in our experiments because of its better theoretical foundation.%

\subsection{Main Results} \label{sec:baseline_results}

The main results of our experiments are summarized in Table~\ref{tab:mega_table_baselines}.

On the BiasBios dataset, our DQN and PPO implementations achieve strong results, with PPO outperforming DQN in fairness, as PPO's lower GAP shows.
 Our supervised implementation, Sup $\mathcal{W}^{EO}$, surpasses all other baselines. LinUCB performs significantly worse on this task, obtaining one of the worst DTO scores. In contrast, on the Emoji dataset, LinUCB achieves one of the best performance-fairness trade-offs, as indicated by the low DTO score, and is only surpassed by BTEO, which has a slightly higher accuracy. 
 On the other hand, DQN obtains a lower accuracy and PPO lower fairness compared to other metrics.

These results suggest that the classical CMAB algorithm LinUCB excels on binary datasets, while the deep RL implementations, DQN and PPO, perform better in multi-class settings.
Notably, although DQN and PPO obtain competitive results on the BiasBios, their F1 score is considerably lower than the baselines. Further analysis of per class metrics (see Appendix~\ref{sec:appendix_recall_perprof}) reveals that while the F1 for most classes was on par with the supervised setup, both deep RL algorithms failed to recall two of the very sparse classes. 

Compared to baseline methods such as BTEO and DAdv, our DQN and PPO implementations demonstrate competitive performance on the BiasBios dataset.
Table \ref{tab:mega_table_baselines} also shows that, contrary to previous work \cite{han2022balancing},\footnote{The EO scaled supervised implementation of \citet{han2022balancing} achieves an Accuracy of 75.7 and GAP of 13.9} 
loss-scaling for supervised learning (Sup $\mathcal{W}^{EO}$) achieves superior overall performance to downsampling (BTEO). Downsampling only seems to outperform scaling when enough data is present, as demonstrated by its lower GAP for the Emoji dataset.

\definecolor{lightblue}{rgb}{0, 0.1, 0.8}
\definecolor{darkred}{rgb}{0.7, 0, 0}

\begin{table}[h]
    \centering
    \resizebox{\columnwidth}{!}{%
    \begin{tabular}{lrrr}
        \toprule
        \multicolumn{1}{c}{\textbf{Algo} + $\mathcal{W}^{EO}$}  & \multicolumn{1}{c}{\textbf{Accuracy} \small{$\uparrow$}}  & \multicolumn{1}{c}{\textbf{GAP} \small{$\downarrow$}}& \multicolumn{1}{c}{\textbf{F1} \small{$\uparrow$}} \\ \midrule
         $\text{Sup}$& \textbf{81.0} (\textcolor{darkred}{- 0.9}) &16.4 (\textcolor{lightblue}{- 9.3}) & \textbf{73.8} (\textcolor{darkred}{- 2.1}) \\       
         $\text{LinUCB}$& 78.4 (\textcolor{darkred}{- 3.8}) & 15.5 (\textcolor{lightblue}{- 3.3}) & 67.3 (\textcolor{darkred}{- 7.5}) \\
         $\text{DQN}_{bandit}$& 80.1 (\textcolor{darkred}{- 0.9}) & \textbf{13.7} (\textcolor{lightblue}{- 3.6}) & 66.5 (\textcolor{darkred}{- 0.1}) \\ 
         $\text{PPO}_{bandit}$& 79.7 (\textcolor{darkred}{- 0.5}) & 14.4 (\textcolor{lightblue}{- 5.9}) & 67.5 (\textcolor{darkred}{- 1.5}) \\
    \bottomrule       
    \end{tabular}}
    \caption{Results on the Bias dataset without reward scaling, presented as mean and difference from the case without EO, where \textcolor{darkred}{red} (worse), \textcolor{lightblue}{blue} (better).}
    \label{tab:eo_ablation_relative}
\end{table}

\subsection{Reward Scaling Impact} \label{sec:influence_rew_scaling}
We investigate the influence of reward scaling on our models by training them with and without scaling. Table \ref{tab:eo_ablation_relative} presents the results on BiasBios as the mean performance without scaling and the change in metrics when EO scaling is applied.

Without reward scaling the three RL algorithms achieve similar accuracy to the supervised approach but at the cost of a lower F1 score. As mentioned above, the RL algorithms fail on two very sparse classes, which explains the drop in GAP and F1. Failing to classify any instances of a class correctly results in a TPR gap of 0 for that class, since the result is "fair" among both genders.

The EO reward scale significantly reduces the GAP of all implementations, at the cost of a slight decrease in Accuracy and F1 for most models.
However, on LinUCB the scaling causes a large performance reduction with only a small GAP reduction, suggesting that scaling hinders the performance more than it improves the fairness.

To inspect the weak effect of reward scaling on LinUCB, we analyze the TPR gap per profession against the gender imbalance for both cases in Figure~\ref{fig:linucb_tpr_gapplot}.  
Without scaling, LinUCB's performance follows a predictable positive correlation with gender imbalance.
For instance, in the Nurse class ($\sim$ 90\% women), the model performs better for the majority group (women), resulting in a positive TPR gap. 
However, after reward scaling this correlation is inverted, causing the model to perform worse for the majority group and better for the minority group. In case of the Nurse class, the model obtains a negative TPR gap.
This suggests LinUCB is oversensitive to scaling on the BiasBios dataset, causing it to overcompensate and penalize the majority group.

\newcommand\imgwidthVar{0.23}   %
\begin{figure}[htbp]
     \centering
     \begin{subfigure}[b]{\imgwidthVar\textwidth}
         \centering
         \includegraphics[width=\textwidth]{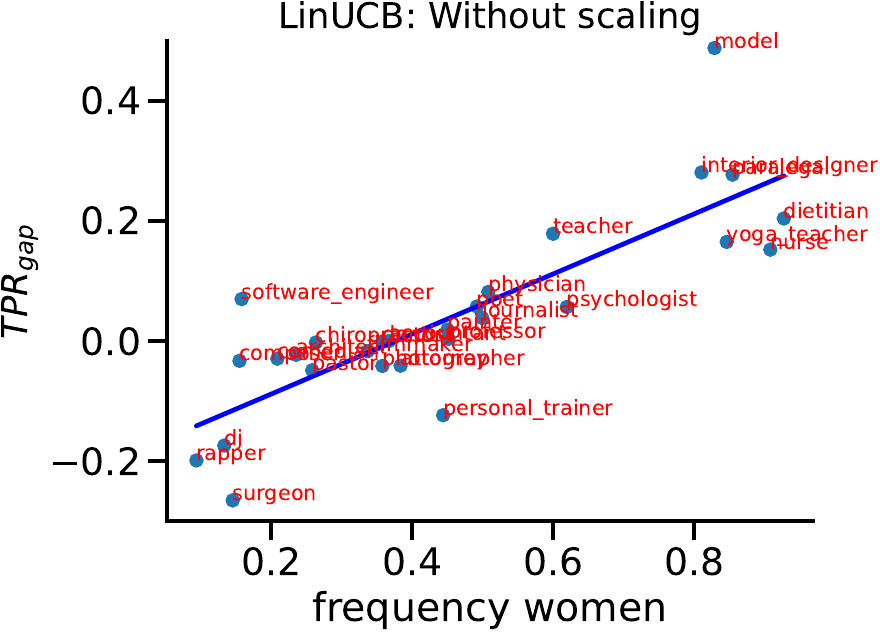}
     \end{subfigure}
     \begin{subfigure}[b]{\imgwidthVar\textwidth}
         \centering
         \includegraphics[width=\textwidth]{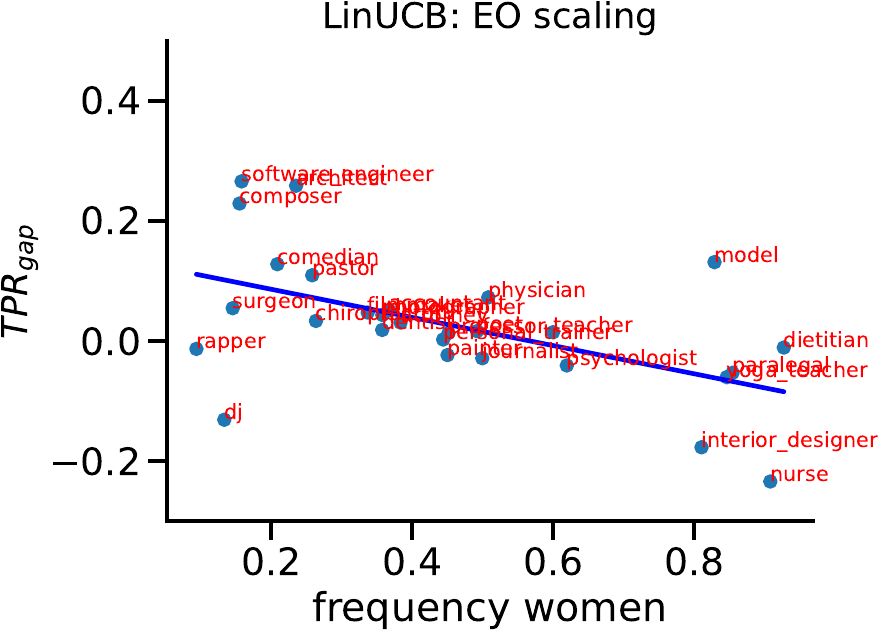}
     \end{subfigure}
        \caption{TPR gap plotted against the gender distribution per profession for LinUCB. Left without reward scaling and right with EO reward scaling}
        \label{fig:linucb_tpr_gapplot}
\end{figure}

\subsection{Subclass Imbalance Sensitivity}

For a more in-depth analysis of each model's sensitivity to subclass imbalance, 
we train them on the Emoji dataset under various stereotyping ratios.
A stereotyping ratio represents the proportion of the AAE and SAE samples in each class. For example, a stereotyping ratio of 0.2 means the data is distributed as Happy (20\% AAE, 80\% SAE), Sad (80\% AAE, 20\% SAE).

\begin{figure}[htbp]
     \centering
     \includegraphics[width=\columnwidth]{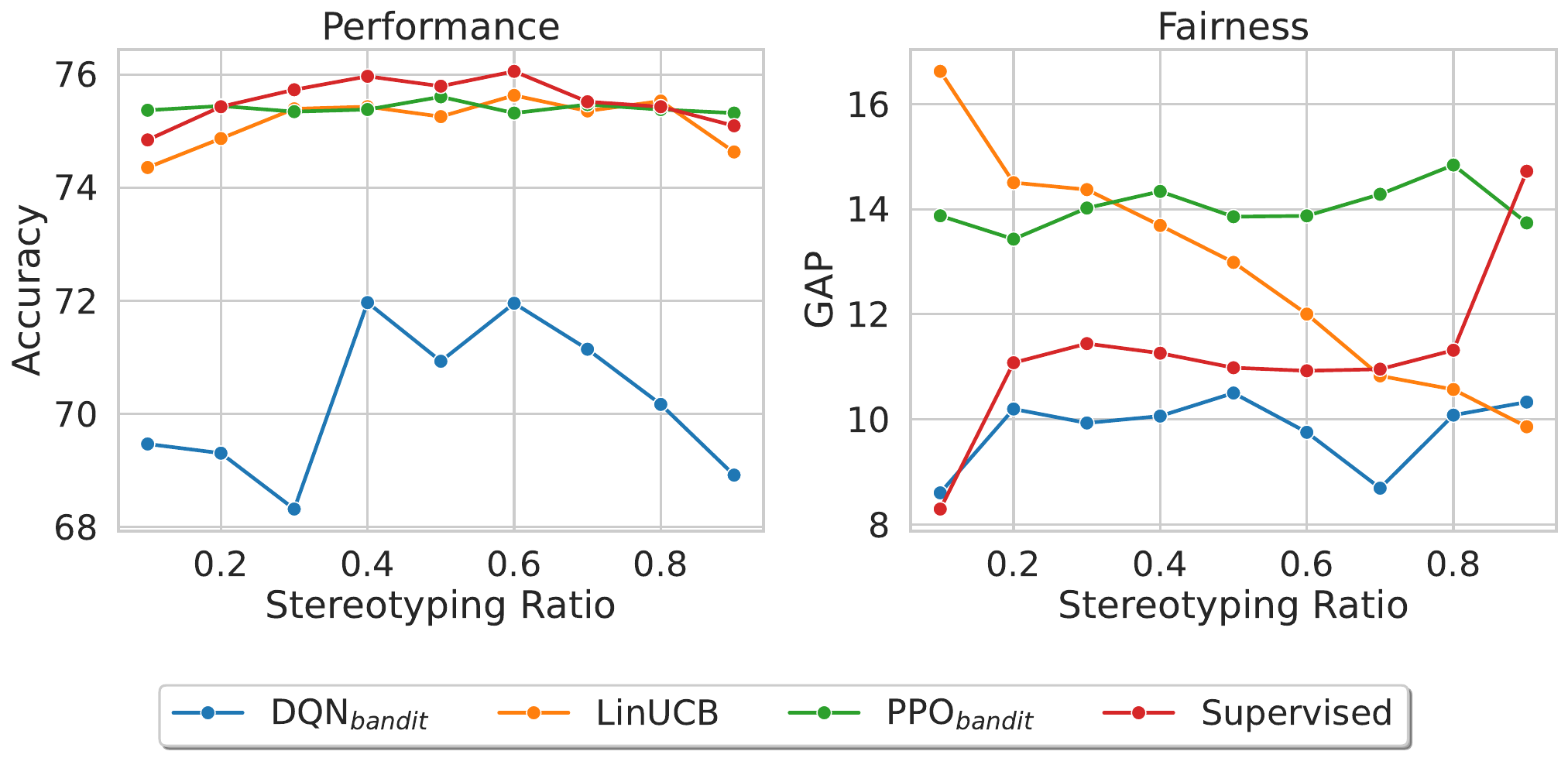}
    \caption{Performance (Accuracy) and Fairness (GAP) on the Emoji dataset using different stereotyping ratios. All models use the scaling of $\mathcal{W}^{EO}$. }
    \label{fig:emoji_ratio}
\end{figure}

Figure \ref{fig:emoji_ratio} reveals a strong inverse relationship between LinUCB's fairness and the stereotyping ratio. Although the stereotypical ratios are symmetric at the value of 0.5 the fairness of LinUCB is asymmetric at this value. 
Thus there is a residual representation bias in the data that is not addressed by the reward scaling. In contrast, supervised learning maintains a mostly stable fairness, except for the most extreme ratios. Interestingly, LinUCB reveals a reverse pattern in best and worst fairness.

The relatively low accuracy of DQN and poor performance on fairness of PPO are consistent across ratios. However, PPO does have the most constant fairness and performance across stereotyping ratios, indicating good training stability.

\subsection{Signal strength vs. Scaling} \label{sec:add_gender_and_mp_debias}
We now examine how the strength of the protected information affects the efficacy of reward scaling.
We focus on two scenarios that modify the gender signal in the representations: 1) adding explicit gender information, thus increasing the gender signal strength 2) debiasing the embeddings using MP, which reduces it. 
Table \ref{tab:genderbool_mpdebias_ablation} presents the mean and relative difference compared to the results without the specified modification.\footnote{Full tables in Appendix \ref{sec:appendix_add_remove_bias}}

Providing the model with gender information increases the overall accuracy. However, the impact on fairness, as indicated by the GAP score varies among algorithms.
The GAP score increases for the two algorithms with the lowest GAPs (Sup, PPO) and decreases for the two with the highest GAP (LinUCB, DQN). 
Only the algorithms that perform less well on fairness benefited from access to protected attribute.

\definecolor{lightblue}{rgb}{0, 0.1, 0.8}
\definecolor{darkred}{rgb}{0.7, 0, 0}

\begin{table}[h]
\centering
\resizebox{\columnwidth}{!}{
\begin{tabular}{@{}lcccc@{}}
        & \multicolumn{2}{c}{Explicit Gender Info} & \multicolumn{2}{c}{MP-Debiased} \\ 
        \cmidrule(lr){2-3} \cmidrule(lr){4-5}
       Algo+$\mathcal{W}^{EO}$ & \textbf{Accuracy} \small{$\uparrow$}& \textbf{GAP} \small{$\downarrow$}& \textbf{Accuracy} \small{$\uparrow$}& \textbf{GAP} \small{$\downarrow$}\\ \midrule
$\text{Sup}$& \textbf{ 80.2} (\textcolor{lightblue}{+ 0.1})& \textbf{7.2} (\textcolor{darkred}{+ 0.1})& \textbf{80.0 }(\textcolor{darkred}{- 0.1})& \textbf{7.4} (\textcolor{darkred}{+ 0.3})\\
$\text{LinUCB}$& 74.5 (\textcolor{lightblue}{+ 0.1})& 11.7 (\textcolor{lightblue}{- 0.5})& 74.3 (\textcolor{darkred}{- 0.3})& 11.5 (\textcolor{lightblue}{- 0.7})\\
$\text{DQN}_{bandit}$& 79.2 (+ 0.0)& 10.0 (\textcolor{lightblue}{- 0.1})& 79.0 (\textcolor{darkred}{- 0.2})& 8.6 (\textcolor{lightblue}{- 1.5})\\ 
$\text{PPO}_{bandit}$& 79.3 (\textcolor{lightblue}{+ 0.1})& 8.7 (\textcolor{darkred}{+ 0.2})& 79.2 (+ 0.0)& 9.7 (\textcolor{darkred}{+ 1.2})\\ \bottomrule
\end{tabular}}
\caption{Results on the BiasBios with added gender info (left) and MP-debiased (right), presented as mean, and difference without change: \textcolor{darkred}{red} (worse), \textcolor{lightblue}{blue} (better).}
\label{tab:genderbool_mpdebias_ablation}
\end{table}

Removing the bias with MP reduces the test accuracy for nearly all algorithms, indicating some useful information is removed.
Again, the modification increased relatively low GAP scores, and decreased relatively high scores. As such, changing to a representation with relatively low bias helps LinUCB and DQN, whereas Sup and PPO that already achieved better fairness mainly see their overall performance hindered.

Notably, the differences in Table \ref{tab:genderbool_mpdebias_ablation} are relatively small and hardly ever surpass the standard deviation provided in Table~\ref{tab:mega_table_baselines}. This suggests that while the strength of the protected information influences performance and fairness, the impact might be less pronounced than the choice algorithmic design. Moreover, all methods reduced the GAP score compared to only applying MP (which yields a GAP of 13.9, see Table~\ref{tab:mega_table_baselines}).

\section{Discussion and Conclusion}
This paper introduces a novel approach to fair classification using the Contextual Multi-Armed Bandit (CMAB) framework and explores various Reinforcement Learning (RL) algorithms. 
Our findings demonstrate the potential of different RL algorithms for this task and the efficacy of reward scaling in mitigating imbalances of protected groups.
The results show the MDP-derived deep RL methods perform best on the multi-class dataset, while the classical bandit algorithm, LinUCB, excels on the binary dataset.
Moreover, our scaled supervised learning implementation achieved new state-of-the-art results on the complex BiasBios dataset.

Our experiments also revealed two limitations 1) RL algorithms can ignore some very sparse classes, despite performing well under most class imbalances.
2) Reward scaling for LinUCB can impair majority group performance beyond that of the minority group.
However, the effect of reward scaling remains robust across varying strengths of the protected information, highlighting its potential as a powerful tool for achieving fair outcomes.

Despite these challenges, we believe that the proposed framework presents a promising approach to leverage RL algorithms for fair classification, opening up new research avenues. We encourage future work to extend upon our framework by exploiting different RL characteristics
, such as model updates for MDP algorithms
based on non-differentiable fairness metrics.

\section*{Limitations}
Important limitations of this work can be divided into two sections: 1) Limitations of the dataset and data requirements of our models 2) Limitations specific to our algorithms and experiments, independent of the data.

\paragraph{Data limitation}
Firstly, all datasets considered in this study used English text, which restricts the analysis and might miss other types of biases related to different linguistic and cultural contexts.
Secondly, the protected groups evaluated in this study simplified to binary labels, which excludes people who do not fall into this category such as non-binary individuals and the multidimensional nature of ethnicity.

Our reward scaling approach also requires these labels for classification.
Although our setup could easily be extended to cases with more labels, it would be interesting to see fair classification with protected attributes as continuous values.  But due to lack of good benchmarks restricts the evaluation of such cases.

\paragraph{Algorithmic limitation}
Firstly, our paper used two deep RL MDP algorithms and one linear classical CMAB agent.
We recognize that while linear agents have a significant focus in the CMAB literature, the fast field of CMAB agents includes options with non-linear algorithms that could also be applied to this task. 
The choice of LinUCB does not represent the state-of-the-art, but rather a classical high-performance implementation.\\
Second, the various hyperparameters limit the extent of general statements about each algorithm. 
We have documented our hyperparameter search and training methods in the appendix, to ensure the interpretability of our experiments, but our results only demonstrate the capabilities of our best implementation. 
Moreover, the use of DTO to select the best model throughout training fails to account for potential trade-offs between fairness and accuracy at different points in training. For example, on the Emoji dataset, PPO underperformed in Fairness and DQN in accuracy. 
However, it is possible that at another pointing training with a higher DTO score, the trade-off between fairness and accuracy was reversed.

\section*{Ethics Statement}
The application of the paper was to improve fairness among protected groups in classification. However, no algorithm is able to obtain perfect fairness and remove the bias perfectly. Therefore applications of the mentioned debiasing methods should always strongly take the mentioned limitations into account. Moreover, the current experiments are limited to specific datasets and real world use cases may be different. Careful evaluation and testing system behavior in the intended setting with input from experts who can judge the consequences of remaining bias is essential.

\section*{Acknowledgments}
This research was partially funded by the Hybrid Intelligence Center, a 10-year program funded by the Dutch Ministry of Education, Culture and Science through the Netherlands Organisation for Scientific Research, \url{https://hybrid-intelligence-centre.nl}.

\bibliography{references}
\bibliographystyle{acl_natbib}

\appendix
\label{sec:appendix}

\section{Reproducibility}
\subsection{Data Analysis}\label{data_analysis}
Because the BiasBios dataset needs to be scraped online, we provide the full composition of the BiasBios dataset split up in profession and gender in Table \ref{tab:full_biasbios_dataset_comp}.

\begin{table}[ht]
    \begin{tabular}{@{}lll@{}}
    \toprule
    \textbf{Profession}   & \textbf{Female} & \textbf{Male} \\ \midrule
    Professor         & 53290           & 64820         \\
    Physician          & 19579          & 18986          \\
    Attorney           & 12494          & 20113          \\
    Photographer       & 8689           & 15635          \\
    Journalist         & 9873           & 10077          \\
    Nurse              & 17236          & 1735           \\
    Psychologist       & 11385          & 6910           \\
    Teacher            & 9768           & 6428           \\
    Dentist            & 5153           & 9326           \\
    Surgeon            & 1972           & 11301          \\
    Architect          & 2398           & 7715           \\
    Painter            & 3543           & 4193           \\
    Model              & 6214           & 1288           \\
    Poet               & 3441           & 3570           \\
    Filmmaker          & 2310           & 4699           \\
    Software Engineer  & 1089           & 5817           \\
    Accountant         & 2081           & 3571           \\
    Composer           & 918            & 4682           \\
    Dietitian          & 3689           & 289            \\
    Comedian           & 592            & 2207           \\
    Chiropractor       & 690            & 1908           \\
    Pastor             & 609            & 1923           \\
    Paralegal          & 1499           & 268            \\
    Yoga Teacher       & 1406           & 257            \\
    Dj                 & 211            & 1274           \\
    Interior Designer  & 1183           & 280            \\
    Personal Trainer   & 654            & 778            \\
    Rapper             & 136            & 1271           \\
    rapper             & 136            & 1271           \\ \bottomrule
    \end{tabular}%
    \caption{Class and gender composition of the BiasBios dataset}
    \label{tab:full_biasbios_dataset_comp}
\end{table}

\subsection{Model Selection} \label{sec:model_selection_dto}
Selecting the best model throughout training or across hyperparameters is strongly dependent on the selection metric. To balance fairness and performance we use the proposed method of \newcite{han2022balancing}, and select using DTO.  The full equation of DTO is provided below,  where the obtained metrics are determined by the point $\left(Acc, (1-GAP)\right)$, and the utopian metrics are  $\left(Acc^{utop}, (1-GAP^{utop})\right)$.

\resizebox{\columnwidth}{!}{
$DTO =\sqrt{ \left(Acc^{utop} - Acc\right)^2 + \left((1-GAP^{utop}) - (1- GAP)\right)^2}$
}

The best training timestep according to DTO is determined with utopian values (1,1), and the best hyperparameters setting utopian values as the best metric values during training (i.e. the highest performance and fairness each individually obtained, which do not necessarily belonging to the same algorithm).

The reported DTO values in table X and Y are obtained using the best performance and accuracy method as: [performance, fairness] BiasBios 28C = [0.811, 0.929], BiasBios 8C = [0.868, 0.978], Moji=[0.756, 0.900]

\subsection{Hyperparameters} \label{sec:hyperparam_tuning}

The architecture of the neural network for each algorithm is fixed and consists of 2 layers MLP. 
For the critic in PPO the architecture is the same except for the output size which is 1.
Hyperparameter optimization is applied for each of the parameters of the algorithms using grid search. Table  \ref{tab:hyperparams_all} shows the ranges and the best values.

\begin{table}[htbp]
    \centering
    \begin{tabular}{lll}
        \toprule
         & Type & Dimensions \\
        \midrule
        Layer 1 & Linear & $n\_features \times 128$ \\
        Layer 2 & Linear & $128 \times n\_actions$ \\
        Activation & ReLU & \\
        Optimizer & Adam & \\

        \bottomrule
    \end{tabular}
    \label{tab:nn_architecture}
    \caption{Neural Network Architecture}

\end{table}

\begin{table*}[t]
    \centering
    \begin{tabular}{llccccc}
        \toprule
        \multirow{2}{*}{Algorithm} & \multirow{2}{*}{Parameter} & \multirow{2}{*}{Min} & \multirow{2}{*}{Max} & \multicolumn{2}{c}{Best} \\
        \cmidrule(lr){5-6}
        & & & & BiasBios & Emoji \\
        \midrule
         & lr (actor) & $3.0 \times 10^{-4}$ & $1.0 \times 10^{-6}$ & $1.0 \times 10^{-4}$ & $3.0 \times 10^{-5}$ \\
         & lr (critic) & $1.0 \times 10^{-3}$ & $1.0 \times 10^{-5}$ & $1.0 \times 10^{-3}$ & $1.0 \times 10^{-4}$ \\
        PPO & Batch size & 64 & 512 & 512 & 512 \\
         & Entropy $c_2$ & 0.01 & 0.1 & 0.2 & 0.1 \\
         & $\epsilon$-clip & 0.05 & 0.3 & 0.1 & 0.3 \\
        \midrule
        Supervised & lr & $1.0 \times 10^{-3}$ & $1.0 \times 10^{-6}$ & $3.0 \times 10^{-4}$ & $1.0 \times 10^{-3}$ \\
         & Batch size & 64 & 512 & 128 & 512 \\
        \midrule
         & lr & $3.0 \times 10^{-4}$ & $1.0 \times 10^{-6}$ & $3.0 \times 10^{-6}$ & $3.0 \times 10^{-4}$ \\
        DQN & Batch size & 32 & 256 & 256 & 32 \\
         & Eps\_end & 0.001 & 0.1 & 0.1 & 0.01 \\
         & Eps decay & 0.5 & 1.0 & 0.5 & 0.5 \\
        \midrule
        LinUCB & $\alpha$  & 0.1 & 3.0 & 1.5 & 2.5 \\
        \bottomrule
    \end{tabular}
    \caption{Hyperparameter ranges and best values for different algorithms. For PPO the "Entropy $c_2$" refers to the coefficient of the entropy in the loss.}
    \label{tab:hyperparams_all}
\end{table*}

\paragraph{Related work implementations}
Following previous work \cite{ravfogel2020null, han2022fairlib}, we use INLP and MP in a post hoc manner to the features extracted from the last hidden layer of the supervised model and train a logistic classifier for the final classification.
For our MP debiasing experiments in section \ref{sec:add_gender_and_mp_debias}
we use MP to debias the context vectors before training, instead of poshoc on the hidden layer of the trained network.

\section{Algorithms}
\subsection{Single-Step Markov Decision Process} \label{single_step_mdp}
To formalize how the policy-gradient methods such as PPO relate to the Contextual Multi-Armed Bandit framework, we define below the single-step Markov Decision Process. An MDP is defined by the tuple $(S, A, P, R, \gamma)$, and our single-step variant contains only two states $S = \{s_1 , s_2\}$. The initial state is sampled each time from the environment and for our classification setup is part of the set of context embeddings, $s_1 \in  \{ x_j \}$. To ensure data samples are treated independently the second state is always the terminal state $s_2 = s_{terminal}$. The action space is equal to the number of classes: $A = C = \{c_1, c_2, .., c_{28}  \}$.  The reward function $R$ is equal to that of the CMAB and is defined in section \ref{sec:cmab}. Lastly, each trajectory is defined as $\tau = \{ s_1, a_1, s_{terminal} \}$ and both the transition probability, $P$,  and the discount factor $\gamma$ are irrelevant since each action results in the terminal state.

\subsection{LinUCB}\label{sec:appendix_linucb}
The full algorithm of LinUCB from \newcite{linucb_li2010contextual}, used in the paper is shown in Algorithm
\begin{algorithm}[ht]
\begin{algorithmic}
    \Require Context features $x_{t,a}$ for context at time $t$ and arm $a \in \mathcal{A}$, exploration parameter $\alpha$. 
    \State Initialize $A_a$ and $b_a$ for each arm $a \in \mathcal{A}$
    \For{each sample $t$}
        \For{each arm $a$}
            \State $\hat{\theta}_{a_t} = A_{a_t}^{-1}b_{a_t}$
            \State  $p_{t,a} = \hat{\theta}_a^\top  x_{t,a}+ \alpha \sqrt{x_{t,a}^\top A_a^{-1} x_{t,a}}$
        \EndFor
        \State Choose arm $a_t = \arg \max_{a \in \mathcal{A}}(p_{t, a})$ , and observe real-valued payoff $r_t$
        \State Update $A_{a_t} \leftarrow A_{a_t}+x_{t, a_t} x_{t, a_t}^{\top}$
        \State Update $b_{a_t} \leftarrow b_{a_t}+r_t x_{t, a_t}$
    \EndFor
    \caption{LinUCB Algorithm}\label{algo:linucb}
\end{algorithmic}
\end{algorithm}

\subsection{Equal Opportunity Weights} \label{sec:eo_appendix}
Where \citet{han2022balancing} used EO for supervised learning, their implementation achieved this objective by grouping the loss per class and then averaging over them. In this section, we see how we can use this to obtain the weights for each data sample based on the class $a$ and protected attribute $g$.
For two protected groups $g_1$ and $g_2$ in class $a$ , let $C_1$ and $C_2$ be the number of samples for $g_1$ and $g_2$,
and $\mathcal{W}_1 $ and $\mathcal{W}_2$, be the weights.
To get a statement of the weights with EO for each sensitive state, $(a,g)$, we need two axioms.

\textbf{Axiom 1.} The weight scale ratio between the two protected groups of a class should be inversely proportional to their probability in the dataset:

$$ \mathcal{W}_1 \cdot C_1 =  \mathcal{W}_2 \cdot C_2 $$

\textbf{Axiom 2:} To ensure fairness across classes, the average weight per profession should be a fixed value $B$ so that:
$$  \frac{1}{ C_1 + C_2 } \left( \mathcal{W}_1 \cdot C_1 + \mathcal{W}_2 \cdot C_2 \right) = B $$

Combining these two axioms we obtain the formulation: 
$$ \mathcal{W}_2 = \frac{B}{2} \frac{( C_1 + C_2)}{C_2} $$

$$ \mathcal{W}_2 = \frac{B}{2} \frac{1}{P(C_2)} $$

For the multi-class classification task the average reward scale, $B$, should be 1, and the probability is conditional on the class $a$, obtaining the final $W_{EO}$ equation:
$$ \mathcal{W}_{EO}(g,y) = \frac{1}{2} \frac{1}{P(g | a)} $$

\section{Ablation Experiments}
Here we add our experiments that did not make the main paper.

\subsection{Analysis: Model and Data Efficiency}
An important aspect for evaluation is related to the data and computational of each algorithm.
For ease of comparison, all algorithms except LinUCB were trained for 10 epochs. However, DQN and PPO each reuse the seen data in a different way to deal with the data sparsity of standard RL settings. DQN is updated using a replay-buffer from which it samples a minibatch of N triplets $(s,a,r)$ for each iteration. In contrast, PPO collects N samples during the observation phase after which it updates the model with this batch $K\_epoch$ number of times. 
Lastly, LinUCB achieves optimal results after 1 epoch but is constrained by the computations of its weight matrices, which require the inverse of a square matrix with dimension $n\_features$. 
For computational efficiency, we use the Sherman–Morrison formula which updates the previous computed inverse with a rank one update \cite{sherman1950adjustment} 

The time complexities in Table \ref{tab:mega_table_baselines}, demonstrate that PPO is closest to supervised learning and that DQN takes significantly more time since it needs to sample from the buffer at each iteration. 
Notably, LinUCB is strongly dependent on the number of classes, reducing its relative efficiency from 32 to 3 times that of Supervised Learning. The bottleneck here is that it needs to compute an upper confidence bound for each class. \\
Another important feature is the sensitivity to hyperparameters. PPO and DQN are sensitive to several hyperparameters that determine the level of its exploration, such as DQN's mini-batch size or exploration parameter, or PPO's entropy and clipping coefficients.
LinUCB is easiest to implement in this regard and does not require any neural network hyperparameters, but only one exploration parameter $\alpha$, see section \ref{sec:appendix_linucb}.

\section{Full result for experiments}
To distinguish the sensitivity of gender imbalance and data-sparsity we also run experiments with a subset of the data, following \newcite{aguirre2023selecting}, and select only the professions that have at least 1000 samples for both genders in the test set, resulting in 8 professions.
\subsection{BiasBios: training performance over time}
\begin{figure}[htbp]
     \centering
     \begin{subfigure}[b]{\imgwidthVar\textwidth}
         \centering
         \includegraphics[width=\columnwidth]{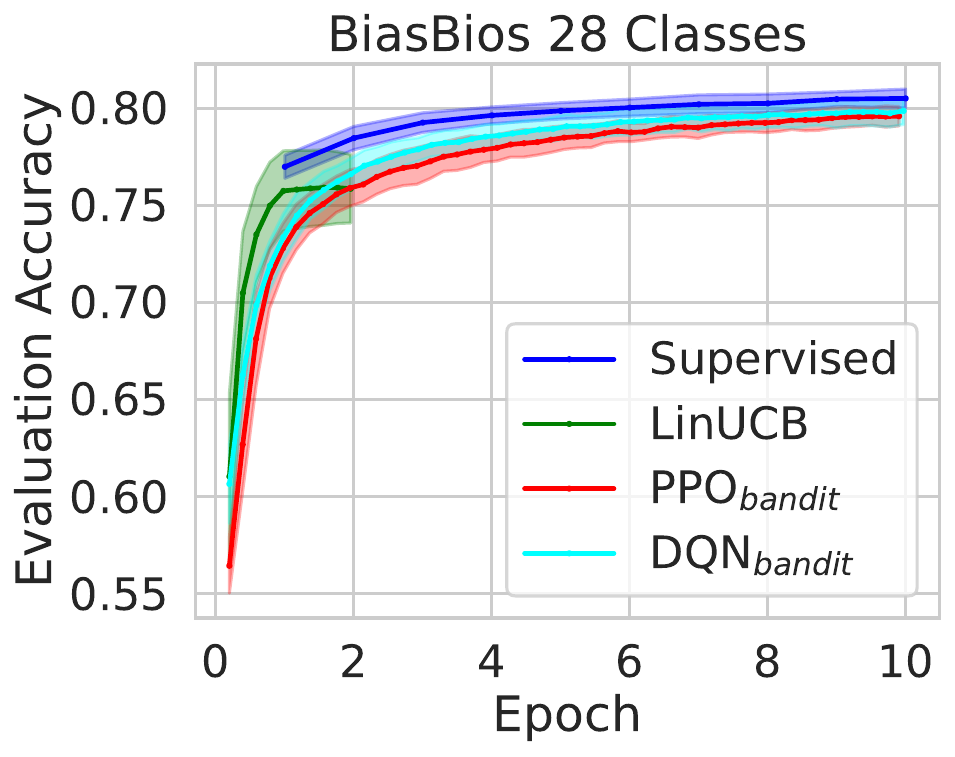}
     \end{subfigure}
     \begin{subfigure}[b]{\imgwidthVar\textwidth}
         \centering
        \includegraphics[width=\columnwidth]{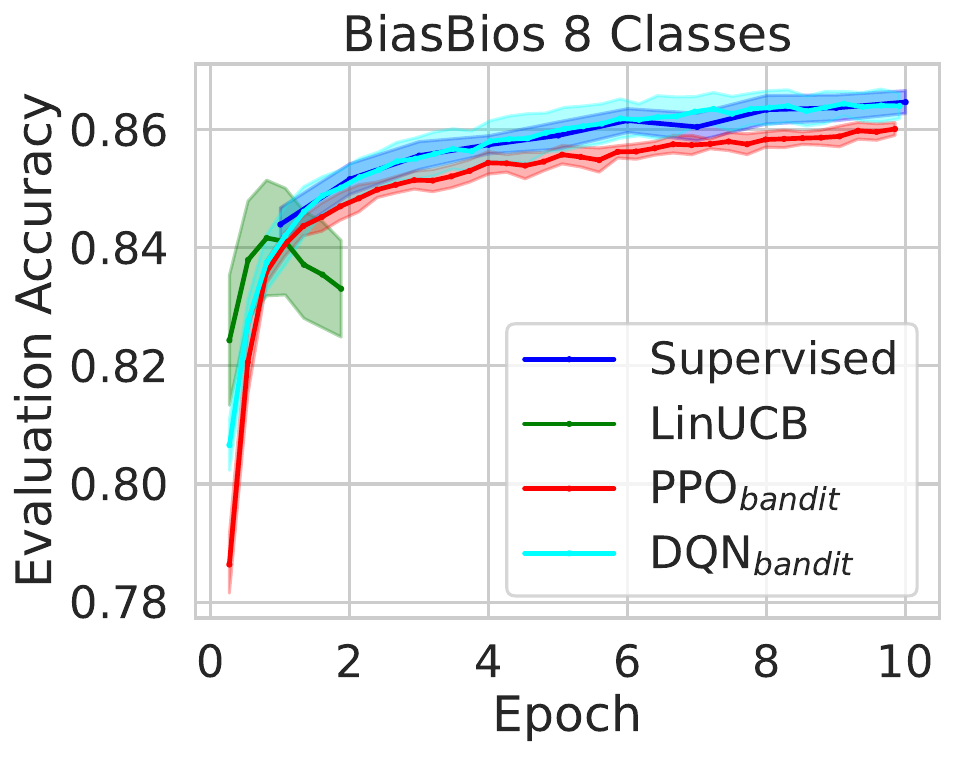}
     \end{subfigure}
    \caption{Evaluation accuracy of the different algorithms the full 28 classes and the 8 class subset of the Bias in Bios dataset}
    \label{fig:train_plot_biasbios}
\end{figure}

In Reinforcement Learning literature it is common to provide the performance of an algorithm throughout training for evaluation. Therefore we provide the evaluation accuracy of our four algorithms in Figure \ref{fig:train_plot_biasbios}

\subsection{BiasBios: Recall per profession} \label{sec:appendix_recall_perprof}
As a further analysis of the lacking F1 score of the RL algorithms compared to the supervised implementation, we provide the Recall scores as a percentage of the class. Since class 21, Professor appears significantly more often than the most common class after it, we leave it out for clarity.

\begin{figure}[htbp]
     \centering
     \begin{subfigure}[b]{\imgwidthVar\textwidth}
         \centering
         \includegraphics[width=\columnwidth]{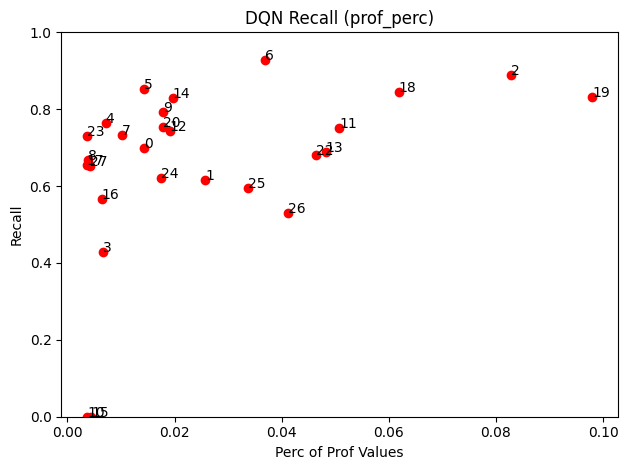}
     \end{subfigure}
     \begin{subfigure}[b]{\imgwidthVar\textwidth}
         \centering
        \includegraphics[width=\columnwidth]{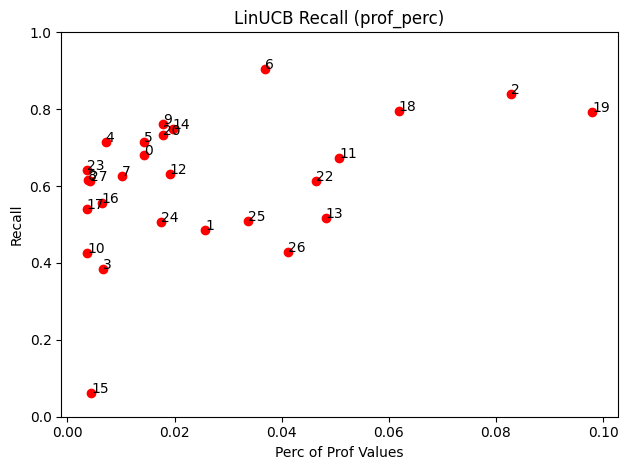}
     \end{subfigure}
     \begin{subfigure}[b]{\imgwidthVar\textwidth}
         \centering
         \includegraphics[width=\columnwidth]{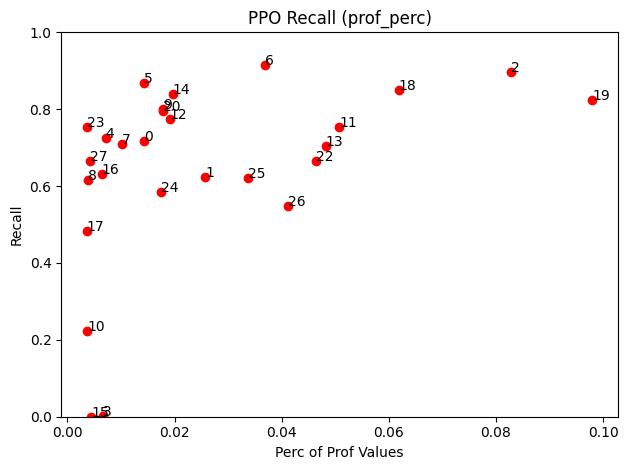}
     \end{subfigure}
     \begin{subfigure}[b]{\imgwidthVar\textwidth}
         \centering
        \includegraphics[width=\columnwidth]{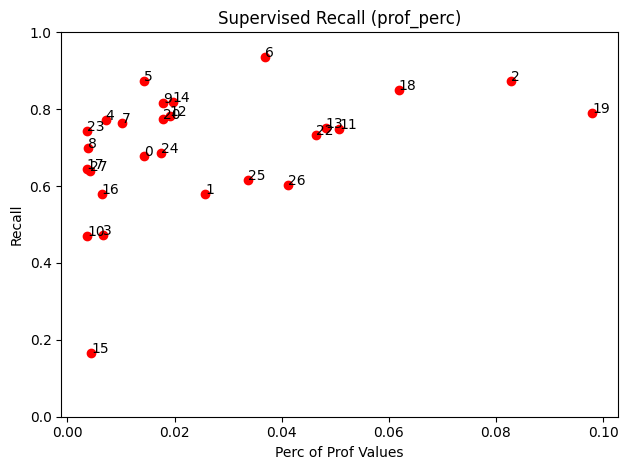}
     \end{subfigure}
    \caption{Recall of each class on the BiasBios dataset for the four algorithm implementations}
    \label{fig:plot_recall_perprof}
\end{figure}

\subsection{Full tables: BiasBios (28C and 8C)}
Some of our results in section \ref{sec:influence_rew_scaling} are presented as the mean only. The full results of our algorithms as the mean and std over the five seeds is provided in the tables here. Table \ref{tab:mega_table_appendix_copy} shows the performance of our algorithms with and without reward scaling on the BiasBios dataset with the 28 and 8 classes.

\begin{table*}[t!]
\centering
\resizebox{\textwidth}{!}{
\begin{tabular}{@{}lcrcccccc}
 & \multicolumn{4}{c}{28 Classes} & \multicolumn{4}{c}{8 Classes} \\
 \cmidrule(lr){2-5} \cmidrule(lr){6-9} 
\multicolumn{1}{c}{\textbf{Algorithm}} & \multicolumn{1}{c}{\textbf{Accuracy} \small{$\uparrow$} } & \multicolumn{1}{c}{\textbf{GAP}\small{$\downarrow$}}  & \multicolumn{1}{c}{\textbf{DTO}}\small{$\downarrow$} & \textbf{F1}\small{$\uparrow$} &  \textbf{Accuracy} \small{$\uparrow$} & \textbf{GAP} \small{$\downarrow$} & \textbf{DTO} \small{$\downarrow$}  & \textbf{F1}\small{$\uparrow$}  \\ \midrule
Sup   & \textbf{81.0 $\pm$ 0.1}& 16.4 $\pm$ 0.5& 10.0& \textbf{73.8 $\pm$ 0.3}&\textbf{ 86.8 $\pm$ 0.1}& 8.3 $\pm$ 0.7& 6.2& \textbf{82.7 $\pm$ 0.1}\\
LinUCB   & 78.4 $\pm$ 0.1& 15.5 $\pm$ 0.3& 9.6& 67.3 $\pm$ 0.4& 85.3 $\pm$ 0.2& \textbf{7.6 $\pm$ 0.3}& 5.8& 80.6 $\pm$ 0.2\\
$\text{DQN}_{bandit}$    & 80.1 $\pm$ 0.2& \textbf{ 13.7 $\pm$ 0.3}&\textbf{7.2}& 66.5 $\pm$ 1.3& 86.5 $\pm$ 0.2& \textbf{7.6 $\pm$ 0.3}&\textbf{ 5.5}& 82.2 $\pm$ 0.2\\
$\text{PPO}_{bandit}$   & 79.7 $\pm$ 0.5& 14.4 $\pm$ 0.7& 8.0& 67.5 $\pm$ 2.0& 86.0 $\pm$ 0.2& 8.7 $\pm$ 0.4& 6.7& 81.6 $\pm$ 0.2\\

\rowcolor{Gray}
Sup$^{ {EO}}$     & \textbf{80.1 $\pm$ 0.2}&\textbf{7.1 $\pm$ 0.5}& \textbf{1.1}& \textbf{ 71.7 $\pm$ 0.5}& \textbf{86.3 $\pm$ 0.2}& 2.4 $\pm$ 0.1& \textbf{0.6}& \textbf{82.0 $\pm$ 0.2}\\
\rowcolor{Gray}
 LinUCB$^{ {EO}}$    & 74.6 $\pm$ 0.2& 12.2 $\pm$ 0.5& 9.6& 59.8 $\pm$1.1& 83.4 $\pm$ 0.2& 7.6 $\pm$ 0.3& 6.8& 77.6 $\pm$ 0.3\\ 
\rowcolor{Gray}
$\text{DQN}_{bandit}^{ {EO}}$   & 79.2 $\pm$ 0.1& 10.1 $\pm$ 0.4& 3.9& 66.4 $\pm$ 0.2& 86.2 $\pm$ 0.1& \textbf{2.2 $\pm$ 0.2}& 0.7& 81.6 $\pm$ 0.2\\
\rowcolor{Gray}
$\text{PPO}_{bandit}^{ ^{EO}}$& 79.2 $\pm$ 0.2& 8.5 $\pm$ 0.2& 2.7& 66.0 $\pm$ 0.8& 85.8 $\pm$ 0.1& 2.8 $\pm$ 0.6& 1.3& 81.4 $\pm$ 0.2\\ \bottomrule
\end{tabular}
}
\caption{Results on the BiasBios dataset for the full dataset (28 classes) and a subset of the most common professions (8 classes). The first rows use a constant reward scale, and the last four (in grey) use the EO reward scale}
\label{tab:mega_table_appendix_copy}
\end{table*}

\subsection{Full tables: four reward scaling methods}\label{sec:appendix_4_4_reward_scales}
The results from reward scaling using the four described scales and our four algorithms are shown in Table \ref{tab:4_4_scales_ablation}.
\begin{table}[htbp]
\centering
\resizebox{\columnwidth}{!}{
\begin{tabular}{@{}llccc} 
\toprule
\textbf{Algo}&       & \textbf{Accuracy} \small{$\uparrow$} & \textbf{GAP} \small{$\downarrow$} & \textbf{F1} \\ \midrule
\multirow{4}{*}{\rotatebox[origin=c]{90}{SUP}} 
                   & $\mathcal{W}_{\rho +}$ & 79.3 $\pm$ 0.1& 7.9 $\pm$ 0.3 & 69.3 $\pm$ 0.3\\
                   & $\mathcal{W}_{\rho -}$ & 79.8 $\pm$ 0.3& 6.9 $\pm$ 0.2 & \textbf{71.8 $\pm$ 0.6} \\
                   & $\mathcal{W}_{EO}$     & \textbf{80.1 $\pm$ 0.2}& 7.1 $\pm$ 0.5 & 71.7 $\pm$ 0.5\\
                   & $\mathcal{W}_{IPW}$    & 72.1 $\pm$ 0.7& \textbf{6.1 $\pm$ 0.3} & 64.8 $\pm$ 0.8\\ \midrule
\multirow{4}{*}{\rotatebox[origin=c]{90}{PPO}} 
                   & $\mathcal{W}_{\rho +}$ & 74.6 $\pm$ 0.7& 9.9 $\pm$ 0.8 & 49.7 $\pm$ 2.2\\
                   & $\mathcal{W}_{\rho -}$ & 78.8 $\pm$ 0.1& \textbf{8.4 $\pm$ 0.6} & 64.7 $\pm$ 0.8\\
                   & $\mathcal{W}_{EO}$     & \textbf{79.2 $\pm$ 0.2}& 8.5 $\pm$ 0.2 & \textbf{66.0 $\pm$ 0.8} \\
                   & $\mathcal{W}_{IPW}$    & 45.8 $\pm$ 6.9& 10.5 $\pm$ 0.9 & 45.3 $\pm$ 5.8\\ \midrule
\multirow{4}{*}{\rotatebox[origin=c]{90}{DQN}} 
                   & $\mathcal{W}_{\rho +}$ & 76.2 $\pm$ 1.1& 10.4 $\pm$ 0.7& 57.2 $\pm$ 4.8\\
                   & $\mathcal{W}_{\rho -}$ & \textbf{79.3 $\pm$ 0.1}& 11.1 $\pm$ 0.6& 65.8 $\pm$ 1.4\\
                   & $\mathcal{W}_{EO}$     & 79.2 $\pm$ 0.1& \textbf{10.1 $\pm$ 0.4}& \textbf{66.4 $\pm$ 0.2} \\
                   & $\mathcal{W}_{IPW}$    & 74.6 $\pm$ 0.3& 12.8 $\pm$ 0.2& 56.6 $\pm$ 0.3\\ \midrule
\multirow{4}{*}{\rotatebox[origin=c]{90}{LinUCB}} 
                   & $\mathcal{W}_{\rho +}$ & 72.8 $\pm$ 0.1& 12.0 $\pm$ 0.5& 54.6 $\pm$ 0.9\\
                   & $\mathcal{W}_{\rho -}$ & 74.1 $\pm$ 0.4& 11.6 $\pm$ 0.5& 59.3 $\pm$ 1.7\\
                   & $\mathcal{W}_{EO}$     & \textbf{74.6 $\pm$ 0.2}& 12.2 $\pm$ 0.5& \textbf{59.8 $\pm$ 1.1} \\
                   & $\mathcal{W}_{IPW}$    & 37.3 $\pm$ 2.5& \textbf{10.3 $\pm$ 0.7}& 35.4 $\pm$ 1.0\\ \bottomrule
\end{tabular}
}
\caption{Results with different reward scaling on BiasBios for various algorithms}
\label{tab:4_4_scales_ablation}
\end{table}

\subsection{Full results: Explicit gender information and Ensemble techniques}\label{sec:appendix_add_remove_bias}
This section includes the full results of Section \ref{sec:add_gender_and_mp_debias},  after adding the gender information explicitely and after removing it with MP. The results are presented as mean and standard deviation over 5 seeds in Table \ref{tab:gender_bool_ablation} and Table \ref{tab:debiased_ablation}
\begin{table}[htbp]
    \centering
    \resizebox{\columnwidth}{!}{%
    \begin{tabular}{lccc}
        \toprule
        \multicolumn{1}{c}{\textbf{Algo + \textit{g}}}  & \multicolumn{1}{c}{\textbf{Accuracy} \small{$\uparrow$}}  & \multicolumn{1}{c}{\textbf{GAP} \small{$\downarrow$}}& \multicolumn{1}{c}{\textbf{F1} \small{$\uparrow$}} \\ \midrule
         $\text{Sup}^{ {EO}}$& \textbf{ 80.2 $\pm$ 0.2}&  \textbf{7.2 $\pm$ 0.5}& \textbf{71.9 $\pm$ 0.7}\\       
         $\text{LinUCB}^{ {EO}}$ &  74.5 $\pm$ 0.2&  11.7 $\pm$ 0.5& 59.6 $\pm$ 0.8\\
         $\text{DQN}_{bandit}^{ {EO}}$&  79.2 $\pm$ 0.2&  10.0 $\pm$ 0.5& 66.1 $\pm$ 0.4\\    
         $\text{PPO}_{bandit}^{ {EO}}$&  79.3 $\pm$ 0.1&  8.7 $\pm$ 0.3& 66.1 $\pm$ 0.6\\  
\bottomrule       
    \end{tabular}}
    \caption{Results on the BiasBios dataset with explicit gender information added to the context.}
    \label{tab:gender_bool_ablation}
\end{table}

\begin{table}[htbp]
    \centering
    \resizebox{\columnwidth}{!}{%
    \begin{tabular}{lccc}
        \toprule
        \multicolumn{1}{c}{\textbf{Algo + \textit{MP}}}  & \multicolumn{1}{c}{\textbf{Accuracy} \small{$\uparrow$}}  & \multicolumn{1}{c}{\textbf{GAP} \small{$\downarrow$}}& \multicolumn{1}{c}{\textbf{F1} \small{$\uparrow$}} \\ \midrule
         $\text{Sup}^{ {EO}}$ &  80.0 $\pm$ 0.2&  7.4 $\pm$ 0.4& 71.9 $\pm$ 0.3\\ 
         $\text{LinUCB}^{ {EO}}$ &  74.3 $\pm$ 0.4&  11.5 $\pm$ 0.1& 59.4 $\pm$ 1.0\\
         $\text{DQN}_{bandit}^{ {EO}}$  &  79.0 $\pm$ 0.2&  8.6 $\pm$ 0.3& 65.8 $\pm$ 0.6\\
          $\text{PPO}_{bandit}^{ {EO}}$  &  79.2 $\pm$ 0.2&  9.7 $\pm$ 0.6& 66.8 $\pm$ 1.4\\
         \bottomrule
    \end{tabular}}
    \caption{Performance on the BiasBios dataset, using MP debiased embeddings}
    \label{tab:debiased_ablation}
\end{table}

\end{document}